\documentclass{bmvc2k}
\usepackage{multirow}
\usepackage{amssymb}

\title{Fine-Grained Visual Classification with Efficient End-to-end Localization}

\addauthor{Harald Hanselmann$^{\text{1,2}}$}{hanselmann@cs.rwth-aachen.de}{1}
\addauthor{Hermann Ney$^{\text{1,2}}$}{ney@cs.rwth-aachen.de}{2}

\addinstitution{
 Human Language Technology and Pattern Recognition Group \\
 RWTH Aachen University\\
 Aachen, Germany
}
\addinstitution{
 AppTek GmbH \\
 Aachen, Germany
}

\runninghead{H.Hanselmann and H.Ney}{Efficient End-to-end Localization}

\def\eg{\emph{e.g}\bmvaOneDot}

\begin{document}

\maketitle

\begin{abstract}
The term fine-grained visual classification (FGVC) refers to classification tasks where the classes are very similar and the classification model needs to be able to find subtle differences to make the correct prediction. State-of-the-art approaches often include a localization step designed to help a classification network by localizing the relevant parts of the input images. However, this usually requires multiple iterations or passes through a full classification network or complex training schedules. In this work we present an efficient localization module that can be fused with a classification network in an end-to-end setup. On the one hand the module is trained by the gradient flowing back from the classification network. On the other hand, two self-supervised loss functions are introduced to increase the localization accuracy. We evaluate the new model on the three benchmark datasets CUB200-2011, Stanford Cars and FGVC-Aircraft and are able to achieve competitive recognition performance.
\end{abstract}
\section{Introduction}
The research area of fine-grained visual classification (FGVC) addresses classification tasks where the different categories are quite similar in appearance and the differences can be very subtle. Such tasks include the categorization of different animal species or car models. State-of-the-art approaches typically rely on a strong convolutional neural network (CNN) as classification or backbone network. This model is then improved with methods that try to make it easier to find the subtle differences between the classes. One such method is localization where the discriminative region of the image to classify is localized and distracting background is discarded. Another advantage of localization is that the scale of the objects is normalized. However, existing approaches for localization usually require inefficient methods to obtain the discriminative regions such as multiple passes through a full classification network or complicated training schedules that prevent end-to-end integration. For this reason we aim to define an efficient localization module that can be integrated and trained in an end-to-end setup. To this end we define a novel, lightweight network to obtain the parameters defining the discriminative region of the image to classify. This module first generates an attention map which is then used to predict the bounding box of the discriminative region. It is trained by the gradients propagated backwards through the end-to-end training, as well as additional self-supervised loss functions. Overall, only the class labels for the training images are needed to train the model.

We evaluate our model on the three standard benchmark datasets CUB200-2011, Stanford Cars and FGVC-Aircraft and are able to report very competitive recognition accuracies.

\subsection{Related work}
There are several approaches that focus on the localization component of a FGVC system. The spatial transformer network (STN) proposed in \cite{jaderberg2015spatial} learns an affine transformation in an end-to-end setting. However, STNs can be difficult to train and in order to be able to estimate the affine parameters a complex network is necessary (in \cite{jaderberg2015spatial} the same network architecture as for the classification network itself is used). This limits the choice of classification networks if GPU memory is limited. 
In the recurrent attention CNN (RA-CNN) \cite{fu2017look} an end-to-end model is presented that recurrently zooms into discriminative regions of the image. Again, complex networks are used to make the decisions about which patches to extract. Additionally, RA-CNN uses a complex training schedule with pre-training and iteratively fixing the weights of sub-networks.
A similar idea has been proposed in \cite{simonelli2018increasingly}. Here multiple classification networks are trained consecutively, where one network uses the output of the previous network to attend to discriminative regions. The final classification decision is found by building an ensemble of the different classification networks.
The trilinear attention sampling network (TASN) \cite{zheng2019looking} uses attention to sample multiple interesting regions in the input image, which are then forwarded through the network again.
In \cite{Hanselmann_2020_WACV} a method was proposed that aims to avoid having to use a full pass through a classification network to be able to localize the relevant image regions. To achieve this, a separate lightweight localization module is trained that learns to predict attention maps that can be used for localization. However, this requires multiple separate training runs instead of a unified end-to-end model.

\section{Efficient end-to-end localization}
\label{sec:e2e}
\begin{figure}[t]
\begin{center}
   \includegraphics[width=1.0\linewidth]{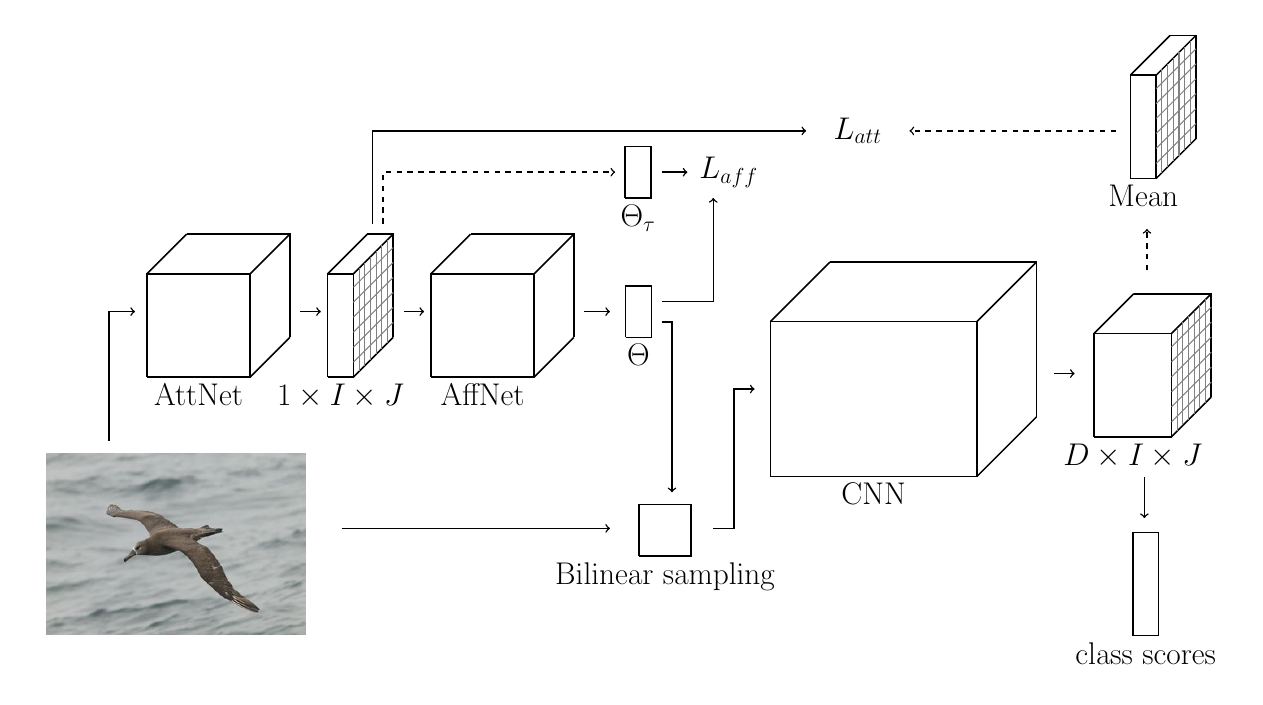}
\end{center}
   \caption{End-to-end model with localization. The dashed arrows indicate that here no gradient flows backwards since these computations are only used for generating the supervision signals for the self-supervised losses.}
\label{fig:e2e_fullsystem}
\end{figure}


Our full end-to-end model consists of three main components, \emph{AttNet}, \emph{AffNet} and the actual classification network. An overview of the system and how the components interact is given in Figure \ref{fig:e2e_fullsystem}. The two components AttNet and AffNet together perform the localization and are designed to be efficient and lightweight such that they do not increase the computation time or memory footprint significantly. Specifically, AttNet generates an attention map of the input image which is then used by AffNet to estimate the affine parameters that define the bounding box of the object in the image. These parameters are then used to return a cropped image which is processed by the classification network. As in the design of STN \cite{jaderberg2015spatial}, the cropping operation is implemented using bilinear sampling. This results in a differentiable cropping procedure which is necessary to train all three components jointly in an end-to-end fashion using gradient descent. 

The full model is trained using four different loss functions, but only a single class label for each training image is needed as annotation. The final output layer of the classification network is trained with the standard cross-entropy loss $L_{CE}$, while a penultimate fully connected layer is trained with the embedding loss $L_{emb}$ as defined in \cite{Hanselmann_2020_WACV}. The latter minimizes the distance of each training sample to its respective class center while maximizing the distances between the class centers. 

In addition to $L_{CE}$ and $L_{emb}$ we define two weakly supervised losses $L_{att}$ and $L_{aff}$ designed to help train AttNet and AffNet. As shown in \cite{zhou2016learning}, the mean $M$ over the feature maps of the final convolutional layer in a classification network defines an attention map of the object in the image. In our setup we use this mean to guide the localization process in two ways. On the one hand we define $L_{att}$ to minimize the distance between the output of AttNet and $M$. This trains AttNet to predict the $M$ for an input image without having to forward the image through the full classification network. On the other hand we calculate bounding box parameters based on $M$ which then serve as target output for AffNet. The difference between this target output and the output of AffNet is then optimized by $L_{aff}$. The exact definition of $L_{att}$ and $L_{aff}$ will be given in Section \ref{section:AttNet} and \ref{section:AffNet}. 

The total overall loss is then given by
\begin{align}
  \label{formula:E2E_overall_loss}
  L = L_{CE} + \lambda L_{emb} + L_{att} + L_{aff}
\end{align}
where the hyper-parameter $\lambda$ is used as a weight for the embedding loss $L_{emb}$.

\subsection{AttNet}
\label{section:AttNet}
Given an input image $X$, AttNet predicts an attention map $A$ of dimension $1 \times I \times J$. Since we do not want to sacrifice computational complexity to obtain $A$, AttNet needs to be lightweight and efficient, but still be able to predict accurate attention maps. It has been shown in \cite{Hanselmann_2020_WACV} that this can be achieved by using the first few layers until after the first residual block of a ResNet-50 \cite{he2016deep} and down-sizing the input to $64 \times 64$. For this reason we chose the same architecture for AttNet.

As part of an end-to-end system AttNet learns from the gradient pushed back from the final layers of the full model (defined by $L_{CE}$ and the embedding loss $L_{emb}$). In addition, we apply local supervision with the loss
\begin{align}
  \label{formula:E2E_LAtt}
  L_{att} &=  \lambda_{att} L_{SL1}(A,M)
\end{align}
where $M$ is defined as the mean over the feature maps of the last convolutional layer and $L_{SL1}$ is the smooth L1 loss \cite{girshick2015fast}.

The influence of $L_{att}$ on the training of AttNet is regulated by the hyper-parameter $\lambda_{att}$. To be able to regulate the influence of the gradient flowing back from the latter part of the model as well we introduce the hyper-parameter $\beta_{att}$. The gradient with respect to $A$ is then computed as

\begin{align}
  \label{formula:E2E_gradient_att}
  \frac{\partial L}{\partial A} &= \beta_{att} \cdot \frac{\partial L_{1}}{\partial A} + \frac{\partial L_{att}}{\partial A} \\
                                &= \beta_{att} \cdot \frac{\partial L_{1}}{\partial A} + \lambda_{att} \cdot \frac{\partial L_{SL1}(A,M)}{\partial A}
\end{align}

where $L_{1}$ is the overall loss without $L_{att}$:
\begin{align}
  \label{formula:E2E_tmp1}
  L_{1} = L_{CE} + \lambda L_{emb} + L_{aff}
\end{align}

By setting $\beta_{att}$ to zero we have the option to cut the gradient flowing back from the classification network and instead train only with the local loss.

\subsection{AffNet}
\label{section:AffNet}
\begin{figure}[t]
\centerline{
\includegraphics[width=0.8\linewidth]{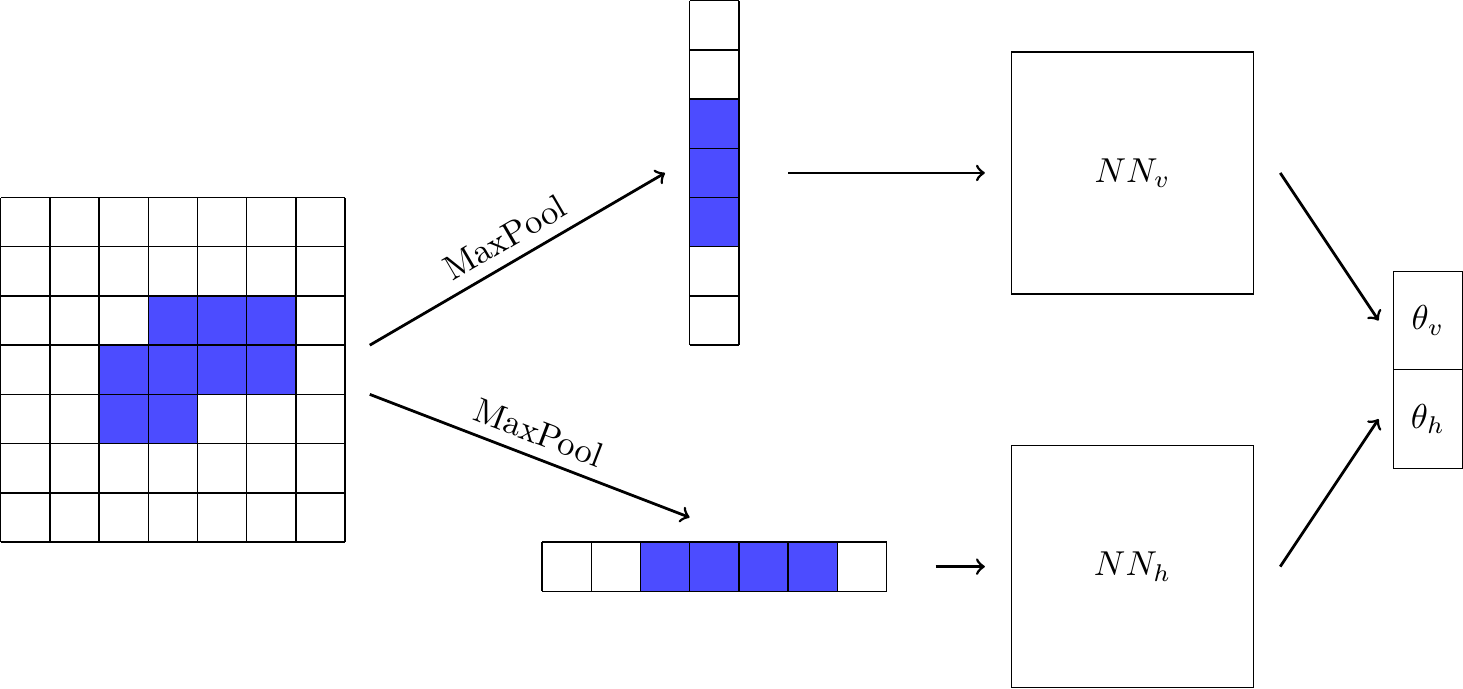}
}
\caption{Illustration of AffNet. Vertical and horizontal max pooling is used to simplify the problem and generate two vectors containing the necessary information to estimate  the vertical and horizontal transformation parameters, respectively.}
\label{fig:EA_affnet}
\end{figure}

Once AttNet has produced an attention map, the task of AffNet is to estimate the affine transformation parameters $\theta$ needed to perform the localization. It operates on single channel input with dimension $1 \times I \times J$ which represents an attention map. AffNet is composed of two parts, a pre-processing module and the affine parameter estimation network.

\subsubsection{Pre-processing module}
The bounding box can be obtained from an attention map by applying min-max normalization and binarization based on a threshold $\tau$ and computing the smallest rectangle containing all positions with value one \cite{Hanselmann_2020_WACV}. In order to incorporate this into AffNet a pre-processing module is defined. However, this module needs to be differentiable in order to be integrated in the end-to-end structure. This can be achieved by defining a network that applies min-max normalization, thresholding and binarization with differentiable layers. Specifically, min-max normalization is achieved by layers computing the minimum and maximum, a subtraction layer and a division layer. The thresholding is realized using a subtraction and a ReLU activation layer. Finally, the binarization is done with a multiplication and Sigmoid activation layer as in suggested \cite{fu2017look}.


Note that the pre-processing module only has one learnable parameter. This is the parameter $w_{\tau}$ that defines the thresholding. It is initialized with the value $\tau$. 

\subsubsection{Affine parameter estimation}
The second component of AffNet is the actual estimation of the affine parameters $\theta$ needed for the localization. Specifically, we define four parameters 
\begin{align}
  \label{formula:theta}
  \theta = [s_x,s_y,t_x,t_y]
\end{align}
where $s_x$ and $s_y$ are the horizontal and vertical scale, while $t_x$ and $t_y$ are the horizontal and vertical translation.

A straightforward way to implement the estimation is to apply some neural network to the attention map and predict the parameters (\eg a simple feed-forward neural network (FFNN) \cite{fu2017look}). However, we have observed that these networks often have difficulty to estimate very accurate parameters. For this reason we split up the problem into a horizontal and vertical part (illustrated in Figure \ref{fig:EA_affnet}). Given an input attention map of size $I \times J$, we apply max pooling with kernels of size $I \times 1$ and $1 \times J$, respectively. Each of the two pooling operations generates a vector containing all necessary information to estimate the scaling and translation parameter for the respective dimension. Given these two vectors two small sub-networks $NN_h$ and $NN_v$ estimate the horizontal and the vertical transformation parameters for translation and scale, respectively. We use the same configuration for both sub-networks. They consist of one linear layer of size 128 with ReLU activation functions followed by a second linear layer of size two that returns the predicted affine parameters for the respective dimension.

Just as with AttNet we also define weak local supervision for AffNet. We generate target transformation parameters $\theta_{\tau}$ from the mean $M$ obtained from the last convolutional layer of the classification network. To achieve this we first compute a bounding box as in \cite{Hanselmann_2020_WACV}. The bounding box coordinates are then converted into the scaling and translation parameters. The loss for AffNet is defined as
\begin{align}
  \label{formula:E2E_LAff}
  L_{aff} =  \lambda_{aff} L_{SL1}(\theta,\theta_{\tau})
\end{align}

Analogously to Formula \ref{formula:E2E_gradient_att} we introduce the hyper-parameter $\beta_{aff}$ to weight the gradient flowing back from the classification and embedding loss. The gradient with respect to $\theta$ is then given by

\begin{align}
  \label{formula:E2E_gradient_aff}
  \frac{\partial L}{\partial \theta} &= \beta_{aff} \cdot \frac{\partial L_{2}}{\partial \theta} + \frac{\partial L_{aff}}{\partial A} \\
                                     &= \beta_{aff} \cdot \frac{\partial L_{2}}{\partial \theta} + \lambda_{aff} \cdot \frac{\partial L_{SL1}(\theta,\theta_{\tau})}{\partial \theta}
\end{align}

where $L_{2}$ is the overall loss without $L_{aff}$:
\begin{align}
  \label{formula:E2E_tmp1}
  L_{2} = L_{CE} + \lambda L_{emb} + L_{att}
\end{align}

\subsubsection{Initialization}
AttNet is created from a pre-trained ResNet-50 and we keep the weights as initialization. AffNet on the other hand is initialized randomly with the exception of the output layer. Here the weights are initialized with zero and the bias is set to one for $s_x$ and $s_y$ and to zero for $t_x$ and $t_y$. As a result, the initial affine transformation is the identity transformation.

\section{Experimental evaluation}
First we evaluate AffNet and compare the model configuration with other possible choices. This is followed by an evaluation of the full end-to-end model.

\subsection{AffNet}
\begin{figure}[h]
\begin{center}
  \begin{tabular} {c c c c c c c c}
    \includegraphics[width=0.09\linewidth]{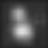} &
    \includegraphics[width=0.09\linewidth]{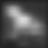} &
    \includegraphics[width=0.09\linewidth]{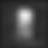} &
    \includegraphics[width=0.09\linewidth]{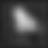} &
    \includegraphics[width=0.09\linewidth]{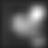} &
    \includegraphics[width=0.09\linewidth]{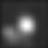} &
    \includegraphics[width=0.09\linewidth]{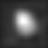} &
    \includegraphics[width=0.09\linewidth]{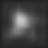} \\
    \includegraphics[width=0.09\linewidth]{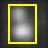} &
    \includegraphics[width=0.09\linewidth]{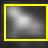} &
    \includegraphics[width=0.09\linewidth]{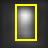} &
    \includegraphics[width=0.09\linewidth]{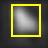} &
    \includegraphics[width=0.09\linewidth]{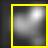} &
    \includegraphics[width=0.09\linewidth]{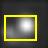} &
    \includegraphics[width=0.09\linewidth]{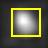} &
    \includegraphics[width=0.09\linewidth]{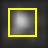} \\
    \includegraphics[width=0.09\linewidth]{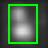} &
    \includegraphics[width=0.09\linewidth]{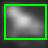} &
    \includegraphics[width=0.09\linewidth]{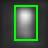} &
    \includegraphics[width=0.09\linewidth]{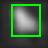} &
    \includegraphics[width=0.09\linewidth]{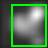} &
    \includegraphics[width=0.09\linewidth]{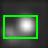} &
    \includegraphics[width=0.09\linewidth]{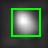} &
    \includegraphics[width=0.09\linewidth]{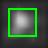} \\
  \end{tabular}
\end{center}
\caption{Example of data used for AffNet evaluation. The upper row contains the input images and the target bounding boxes are shown in the middle row. The lower row shows the actual predictions estimated by AffNet.}
\label{fig:affnet_data}
\end{figure}

We evaluate the design of AffNet with a set of attention maps generated from the CUB200-2011 dataset using a trained classification network without localization. Some examples of the attention maps as well as the target bounding boxes are shown in the first two rows of Figure \ref{fig:affnet_data}. The generated attention maps are split into a training and a test set.
The performance is measured by the smooth L1 error achieved on the test set as well as the mean IoU for two different thresholds. Additionally, we report the runtime in milliseconds.
We compare AffNet with three other architectures which are defined as follows:
\begin{itemize}
 \item \textbf{FFNN}: This model is a small feed-forward network designed to have a similar number of parameters as AffNet. It has one fully connected layer of size 32 with batch normalization and ReLU activation functions.
 \item \textbf{ResNet-S}: In the definition of AttNet we use the first few layers of a ResNet-50 including the first residual module. The same can be done for AffNet and we refer to this shortened model as ResNet-S.
 \item \textbf{ResNet-50}: We also evaluate using a full ResNet-50. To fit the input dimension of $224\times224\times3$ we re-scale the attention maps.
\end{itemize}

All three architectures use an output layer of size four to predict the four affine parameters for scale and translation. We test all architectures with and without applying the pre-processing module first.

The results are given in Tables \ref{table:EA_affnet_results}. Comparing the models with and without the pre-processing module it becomes evident that the pre-processing module helps to improve the predictions significantly. For all network configurations the results for the IoU metric are better by including the pre-processing module. Additionally, we can observe that AffNet achieves the best result with the least parameters and close to the fastest runtime. The second best result with respect to IoU is achieved by the shortened ResNet that was also used as architecture for AttNet. It is on par with AffNet for IoU $> 0.8$, but if the threshold is set to $0.95$ AffNet still achieves decent accuracy, while for ResNet-short the accuracy drops below 10\%. Additionally, AffNet is much faster and needs much fewer parameters. 
The accuracy of AffNet is further illustrated in Figure \ref{fig:affnet_data}. The third row contains predictions made by AffNet and we can observe that they almost match the target bounding boxes.
For these reasons we believe AffNet is the ideal choice to estimate the affine parameters in the localization module for the end-to-end classification model.
\begin{table}
  \begin{center}
    \begin{tabular}{|l|c|c|c|c|c|c|}
      \hline
      \multirow{2}{*}{Network} & Prep.   &\multirow{2}{*}{Parameters} & Runtime & SL1 error               & \multirow{2}{*}{IoU $> 0.8$} & \multirow{2}{*}{IoU $> 0.95$} \\
                               & module &                         &     (ms)                     & ($\times e^{-3}$) &                & \\
      \hline
      \hline
      FFNN         & & \phantom{0000}6500     & \phantom{0}0.1  & 6.4 & 0.46 & 0.00 \\
      FFNN        & \checkmark &\phantom{0000}6501     & \phantom{0}0.3  & 4.3 & 0.61 & 0.01 \\
      \hline
      ResNet-S & & \phantom{00}241732   & \phantom{0}0.9  & 2.7 & 0.80 & 0.01 \\
      ResNet-S & \checkmark &\phantom{00}241733   & \phantom{0}1.0  & 1.1 & 0.97 & 0.09 \\
      \hline
      ResNet-50    & & 23516228 & 28.6 & 5.4 & 0.47 & 0.01 \\ 
      ResNet-50    & \checkmark &23516229 & 30.4 & 1.9 & 0.88 & 0.02 \\
      \hline
      \hline
      AffNet       & & \phantom{0000}4868     & \phantom{0}0.2  & 2.7 & 0.81 & 0.02 \\
      AffNet       &\checkmark & \phantom{0000}4869     & \phantom{0}0.4  & 0.5 & 0.98 & 0.54 \\
      \hline
    \end{tabular}
  \end{center}
  \caption{Comparison of different model architectures to estimate the affine parameters. \label{table:EA_affnet_results}}
\end{table}

\subsection{Full system}
We evaluate the full end-to-end model on the three standard benchmark datasets CUB200-2011 \cite{birds}, Stanford cars \cite{cars} and FGVC-Aircraft \cite{aircraft}. The CUB200-2011 dataset contains 5994 training and 5794 test images for 200 different bird species. The Stanford cars dataset contains 8144 training and 8041 test images for 196 different car models. The FGVC-Aircraft dataset contains 6667 training and 3333 test images for 100 different airplane models. We use the Torch7 framework \cite{collobert2011torch7} to implement and train our models and the code will be made publicly available.

Our training setup is very similar to \cite{Hanselmann_2020_WACV}. We use ResNet-101 \cite{he2016deep} as classification network which has been pre-trained on ImageNet \cite{he2016deep} with overlapping test images removed. The input resolution is $448 \times 448$, and we train for 90 epochs with a starting learning rate of $0.003$. The learning rate gets reduced every 30 epochs by multiplying with 0.1. Apart from the ablation study we set the hyper-parameters $\beta_{att}$ and $\beta_{aff}$ to one and $\lambda_{att}$ and $\lambda_{aff}$ to 16. The threshold $\tau$ that also initializes $w_{\tau}$ is set to 0.3.

We evaluate with two settings. Setting 1 is similar to \cite{Hanselmann_2020_WACV} and we use ResNet-101 with an additional embedding layer with dimension 512. In setting 2 we modify the stem of ResNet-101 as in \cite{he2019bag} (ResNet-C) and use a feature dimension of 1024 for the embedding layer.

Due to the lightweight configuration of AttNet and AffNet, the full model can be trained on a single GPU with 11 GB memory with a batch-size of 14. While a forward pass of the classification network has a runtime of 162 ms, adding the localization module increases this only to 164 ms.


\subsubsection{Ablation study}
In Table \ref{table:e2e_birds} we analyze the importance of the loss $L_{att}$ and $L_{aff}$ as well as the two hyper-parameters $\beta_{att}$ and $\beta_{aff}$. As baseline, we set the latter to zero and disable $L_{att}$ and $L_{aff}$ (by setting $\lambda_{att}$ and $\lambda_{aff}$ to zero as well). This means no local self-supervision is used and no gradient is flowing back through AffNet and AttNet. As a result, the initial parameters are never updated and the localization module is fixed to return the identity transformation. The baseline result of 87.0\% accuracy therefore corresponds to using no localization at all. 
If we keep the two losses disabled but set $\beta_{att}$ and $\beta_{aff}$ to one we end up with a setup similar to STNs \cite{jaderberg2015spatial} where the affine parameters are learned only through the gradient flowing back from the classification loss. However, we can observe that this does not lead to an improvement over the baseline. Using only the local self-supervision and cutting off the gradients by setting $\beta_{att}$ and $\beta_{aff}$ to zero does lead to an improvement over the baseline. This indicates that the local self-supervision is very important to achieve a good recognition performance. Setting $\beta_{att}$ and $\beta_{aff}$ to one then results in a true end-to-end system and yields another improvement with a classification accuracy of 88.5\%.

\begin{table}
  \begin{center}
  \begin{tabular} {|l|c|c|c|c|}
    \hline
    Description & $L_{att}$ and $L_{aff}$ & $\beta_{att}$ & $\beta_{aff}$ & Accuracy[\%] \\
    \hline
    \hline
    Baseline       &  &  &  & 87.0 \\
    \hline
    No local supervision &  & \checkmark & \checkmark & 86.9 \\
    No end-to-end   & \checkmark &  &  & 88.1 \\
    No gradient from AffNet to AttNet & \checkmark & & \checkmark & 88.4 \\
    \hline
    End-to-end & \checkmark & \checkmark & \checkmark & 88.5 \\
    \hline
  \end{tabular}
  \end{center}
  \caption{Ablation study using the CUB200-2011 dataset and setting 1. \label{table:e2e_birds}}
\end{table}

\subsubsection{Comparison to state-of-the-art}
In Table \ref{table:e2e_sota} we compare the end-to-end system presented in this work with the best state-of-the-art results on the CUB200-2011, Stanford Cars and FGVC Aircraft benchmarks. Compared with other methods focusing on localization (\eg \cite{fu2017look,simonelli2018increasingly,Hanselmann_2020_WACV}) we can achieve the best accuracies while also offering a unified and efficient training process on top of efficient testing. On Stanford Cars only TResNet \cite{ridnik2020tresnet} achieves a better result with respect to accuracy. TResNet is a very recently proposed new classification network that could also well be incorporated into our approach by replacing ResNet-101. Also on FGVC Aircraft there is with TBMSL-Net \cite{zhang2020three} only one method with a better accuracy. However, TBMSL-Net is a multi-scale and multi-patch approach, while our model only uses a single pass.

\begin{table}
  \begin{center}
  \begin{tabular} {|l|c|c|c|}
    \hline
    Method  & \multicolumn{3}{c|}{Accuracy[\%]} \\
    \cline{2-4}
            &  CUB200-2011 & Stanford Cars & FGVC Aircraft  \\
    \hline
    \hline
    STN \cite{jaderberg2015spatial}      & 84.1 & -    & - \\
    RA-CNN \cite{fu2017look}             & 85.3 & 92.5 & - \\
    ISE \cite{simonelli2018increasingly} & 87.2 & 94.1 & 90.9 \\
    NTS-Net \cite{yang2018learning}      & 87.5 & 93.9 & 91.4 \\
    DCL \cite{Chen_2019_CVPR}            & 87.8 & 94.5 & 93.0 \\
    OSME-MAMC \cite{sun2018multi}        & 86.5 & 93.0 & - \\
    iSQRT-COV \cite{li2018towards}       & 88.7 & 93.3 & 91.4 \\
    CS Parts \cite{korsch2019class}      & 89.5 & 92.5 & - \\
    Spatial RNN \cite{wu2018deep}        & 89.7 & 93.4 & 88.4 \\
    Stacked LSTM \cite{Ge_2019_CVPR}     & \bf{90.4} & - & - \\
    GPipe \cite{huang2018gpipe}          & -    & 94.8 & 92.9 \\
    AutoAugm \cite{cubuk2018autoaugment} & -    & 94.8 & 92.7 \\
    TResNet \cite{ridnik2020tresnet}     & -    & \bf{96.0} & - \\
    ELoPE \cite{Hanselmann_2020_WACV}    & 88.5 & 95.0 & 93.5 \\
    API-Net \cite{zhuang2020learning}    & 90.0 & 95.3 & 93.9 \\
    TBMSL-Net \cite{zhang2020three}      & 89.6 & 94.7 & \bf{94.5} \\
    \hline
    Ours (setting 1)     & 88.5 & 95.3 & 93.9 \\
    Ours (setting 2)     & 88.9 & 95.6 & 94.1 \\
    \hline
  \end{tabular}
  \end{center}
  \caption{Comparison with state-of-the-art on the CUB200-2011, Stanford Cars and FGVC Aircraft benchmarks. \label{table:e2e_sota}}

\end{table}

\section{Conclusion}
In this work we introduced an efficient localization module that can be integrated into an FGVC model in an end-to-end setup. We presented a new network to estimate affine parameters. This network is composed of two parts, AttNet and AffNet. We showed that our choice for AffNet is able to estimate very accurate parameters given an attention map as input. The localization module achieves the best performance boost when it is trained end-to-end with additional self-supervised loss functions. The latter derive their supervision signals from the mean over the feature maps of the last convolutional layer of the classification network. Especially on Stanford Cars and FGVC Aircraft very competitive recognition accuracies were obtained.

For future work it would be interesting to test other backbone classification networks such as the recently proposed TResNet. This could lead to another boost in classification accuracy.

\bibliography{e2e_loc}
\end{document}